\documentclass[10pt,twocolumn,letterpaper]{article}

\usepackage{cvpr}
\usepackage{times}
\usepackage{epsfig}
\usepackage{graphicx}
\usepackage{amsmath}
\usepackage{amssymb}

\usepackage{amsthm}
\usepackage{mathrsfs}
\usepackage{rotating}
\usepackage{algorithm,algorithmic}
\usepackage{multirow,booktabs}

\newenvironment{myitemize}[1][]{
\begin{list}{{\hei #1}} %
    {
     \setlength{\leftmargin}{1.2em}
     \setlength{\topsep}{0.3em}
     \setlength{\itemsep}{0em}
    }}
{\end{list}}

\newcommand{\PHASE}{\item[\phase]}
\newcommand{\phase}{\vspace{5em}\emph{Phase}}

\newcommand{\MYREPEATE}{\item[\myrepeat]}
\newcommand{\myrepeat}{\textbf{Repeat}}

\newcommand{\MYUNTIL}{\item[\myuntil]}
\newcommand{\myuntil}{\textbf{Until}}

\usepackage[pagebackref=true,breaklinks=true,letterpaper=true,colorlinks,bookmarks=false]{hyperref}

\cvprfinalcopy 

\ifcvprfinal\fi
\begin{document}

\title{Clothing Co-Parsing by Joint Image Segmentation and Labeling}

\author{Wei Yang$^1$, ~ Ping Luo$^2$, ~ Liang Lin$^{1,3}$\thanks{Corresponding author is Liang Lin. This work was supported by the Hi-Tech Research and Development Program of China (no.2013AA013801), Guangdong Natural Science Foundation (no.S2013050014548), Program of Guangzhou Zhujiang Star of Science and Technology (no.2013J2200067), Special Project on Integration of Industry, Educationand Research of Guangdong (no.2012B091100148), and Fundamental Research Funds for the Central Universities (no.13lgjc26).} \\
$^1$Sun Yat-sen University, Guangzhou, China \\
$^2$Department of Information Engineering, The Chinese University of Hong Kong\\
$^3$SYSU-CMU Shunde International Joint Research Institute, Shunde, China\\
{\tt\small \{platero.yang, pluo.lhi\}@gmail.com, linliang@ieee.org}}

\maketitle

\begin{abstract}
This paper aims at developing an integrated system of clothing co-parsing, in order to jointly parse a set of clothing images (unsegmented but annotated with tags) into semantic configurations. We propose a data-driven framework consisting of two phases of inference. The first phase, referred as ``image co-segmentation'', iterates to extract consistent regions on images and jointly refines the regions over all images by employing the exemplar-SVM (E-SVM) technique~\cite{ESVM}. In the second phase (i.e. ``region co-labeling''), we construct a multi-image graphical model by taking the segmented regions as vertices, and incorporate several contexts of clothing configuration (\eg, item location and mutual interactions). The joint label assignment can be solved using the efficient Graph Cuts algorithm. In addition to evaluate our framework on the Fashionista dataset \cite{Fashion}, we construct a dataset called CCP consisting of 2098 high-resolution street fashion photos to demonstrate the performance of our system. We achieve 90.29\% / 88.23\% segmentation accuracy and 65.52\% / 63.89\% recognition rate on the Fashionista and the CCP datasets, respectively, which are superior compared with state-of-the-art methods.
\end{abstract}


\section{Introduction}\label{sec:intro}

\begin{figure*}[t]
\begin{center}
    \epsfig{figure=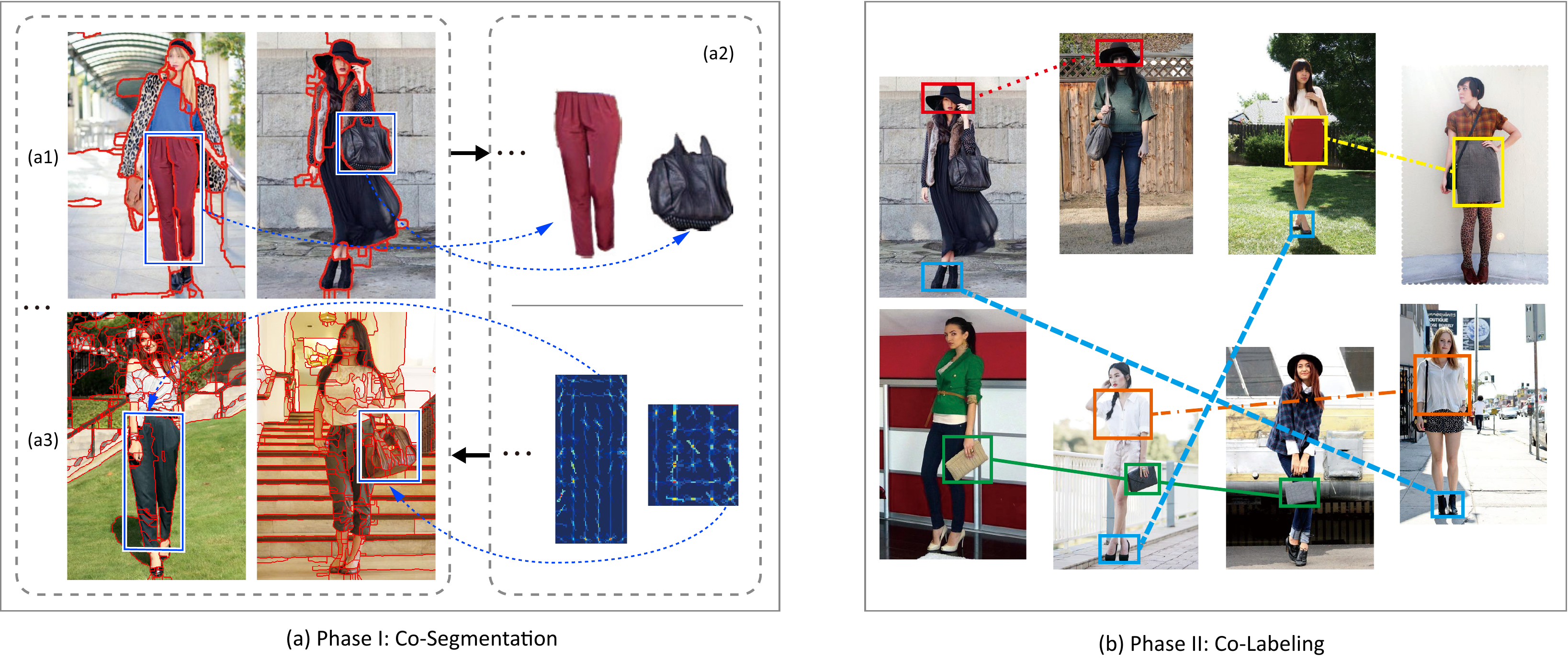, width=1\textwidth}
\end{center}
\vspace{0mm}
   \caption{Illustration of the proposed clothing co-parsing framework, which consists of two sequential phases of optimization: (a) clothing co-segmentation for extracting coherent clothes regions, and (b) region co-labeling for recognizing various clothes garments. Specifically, clothing co-segmentation iterates with three steps: (a1) grouping superpixels into regions, (a2) selecting confident foreground regions to train  E-SVM classifiers, and (a3) propagating segmentations by applying E-SVM templates over all images. Given the segmented regions,  clothing co-labeling is achieved based on a multi-image graphical model, as illustrated in (b).
}\vspace{0mm}
   \label{fig:framework}
\end{figure*}

Clothing recognition and retrieval have huge potentials in internet-based e-commerce, as the revenue of online clothing sale keeps highly increasing every year. In computer vision, several interesting works~\cite{Fashion, ATTRIBUTE, STREET, CLOSET} have been proposed on this task and showed promising results. On one hand, pixelwise labeling of clothing items within images is one of the key resources for the above researches, but it often costs expensively and processes inefficiently. On the other hand, it is feasible to acquire image-level clothing tags based on rich online user data. Therefore, an interesting problem arises, which is the focus of this paper: How to jointly segment the clothing images into regions of clothes and simultaneously transfer semantic tags at image level to these regions.

The key contribution of this paper is an engineered and applicable system\footnote{http://vision.sysu.edu.cn/projects/clothing-co-parsing/} to jointly parse a batch of clothing images and produce accurate pixelwise annotation of clothing items. We consider the following challenges to build such a system:

\begin{myitemize}
    \item[$\bullet$] The appearances of clothes and garment items are often diverse with different styles and textures, compared with other common objects. It is usually hard to segment and recognize clothes via only bottom-up image features.
    \item[$\bullet$] The variations of human poses and self-occlusions are non-trivial issues for clothing recognition, although the clothing images can be in clear resolution and nearly frontal view.
    \item[$\bullet$] The number of fine-grained clothes categories is very large, \eg, more than 50 categories in the Fashionista dataset \cite{Fashion}; the categories are relatively fewer in existing co-segmentation systems \cite{KWAYCOSEG, RECPEOPLE}.
\end{myitemize}

To address the above issues, we develop the system consisting of two sequential phases of inference over a set of clothing images, \ie image co-segmentation for extracting distinguishable clothes regions, and region co-labeling for recognizing various garment items, as illustrated in Figure~\ref{fig:framework}. Furthermore, we exploit contexts of clothing configuration, \eg, spatial locations and mutual relations of clothes items, inspired by the successes of object/scene context modeling \cite{lingrammar,liu2011integrating,lin2012object}.

In the phase of image co-segmentation, the algorithm iteratively refines the regions grouped over all images by employing the exemplar-SVM (E-SVM) technique~\cite{ESVM}. First, we extract superpixels and group them into regions for each image, where most regions are often cluttered and meaningless due to the diversity of clothing appearances and human variations, as shown in Figure~\ref{fig:framework} (a1). Nevertheless, some coherent regions (in Figure~\ref{fig:framework} (a2)) can be still selected that satisfy certain criteria (\eg, size and location).  Then we train a number of E-SVM classifiers for the selected regions using the HOG feature, \ie, one classifier for one region, and produce a set of region-based detectors, as shown in Figure~\ref{fig:framework} (a3), which are applied as top-down templates to localize similar regions over all images. In this way, segmentations are refined jointly, as more coherent regions are generated by the trained E-SVM classifiers. This process is inspired by the observation that clothing items of the same fine-grained category often share similar patterns (\ie shapes and structures). In the literature, Kuettel et al.~\cite{kuettel2012segmentation} also proposed to propagate segmentations through HOG-based matching.




Given the segmented regions of all images, it is very difficult to recognize them by only adopting supervised learning due to the large number of fine-grained categories and the large intra-class variance. In contrast, we perform the second phase of co-labeling in a data-driven manner. We construct a multi-image graphical model by taking the regions as vertices of graph, inspired by \cite{MIM}. In our model, we link adjacent regions within each image as well as regions across different images, which shares similar appearance and latent semantic tags. Thus we can borrow statistical strength from similar regions in different images and assign labels jointly, as Figure~\ref{fig:framework} (b) illustrates. The optimization of co-labeling is solved by the efficient Graph Cuts algorithm~\cite{boykov2001fast} that incorporates several constraints defined upon the clothing contexts.

%
%


Moreover, a new database with groundtruths is proposed for evaluating clothing co-parsing,  including more realistic and general challenges, \eg. disordered backgrounds and multiform human poses, compared with the existing clothes datasets \cite{borras2003high, chen2006cloth, Fashion}. We demonstrate promising performances and applicable potentials of our system in the experiments.

\subsection{Related Work}\label{sec:related-work}

In literature, existing efforts on clothing/human segmentation and recognition mainly focused on constructing expressive  models to address various clothing styles and appearances~\cite{chen2006cloth, hasan2010coseg, bo2011cloth, wang2011coseg,  luo2012hierarchical, Fashion, luo2013pedestrian}. One classic work \cite{chen2006cloth} proposed a composite And-Or graph template for modeling and parsing clothing configurations. Later works studied on blocking models to segment clothes for highly occluded group images \cite{wang2011coseg}, or deformable spatial priors modeling for improving performance of clothing segmentation \cite{hasan2010coseg}.  Recent approaches incorporated shape-based human model \cite{bo2011cloth}, or pose estimation and supervised region labeling \cite{Fashion}, and achieved impressive results.
Despite acknowledged successes, these works have not yet been extended to the problem of clothing co-parsing, and they often require much labeling workload.

Clothing co-parsing is also highly related to image/object co-labeling, where a batch of input images containing similar objects are processed jointly~\cite{liu2009label,luo2012joint,lin2010layered}. For example, unsupervised shape guided approaches were adopted in \cite{leibe2004label} to achieve single object category co-labeling. Winn et. al. \cite{winn2005object} incoporated automatic image segmentation and spatially coherent latent topic model to obtain unsupervised multi-class image labeling. These methods, however, solved the problem in an unsupervised manner, and might be intractable under circumstances with large numbers of categories and diverse appearances.  To deal with more complex scenario, some recent works focused on supervised label propagation,  utilizing pixelwise label map in the training set and propagating labels to unseen images. Pioneering work of Liu et al. \cite{liu2009label} proposed to propagate labels over scene images using a bi-layer sparse coding formulation. Similar ideas were also explored in \cite{liu2011label}. These methods, however,  are often limited by expensive annotations. In addition, they extracted image correspondences upon the pixels (or superpixels), which are not discriminative for the clothing parsing problem.


The rest of this paper is organized as follows. We first introduce the probabilistic formulation of our framework in Section~\ref{sec:formulation}, and then discuss the implementation of the two phases in Section ~\ref{sec:co-parsing}. The experiments and comparisons are presented in Section~\ref{sec:experiments}, and finally comes the conclusion in Section~\ref{sec:conclusions}.

\section{Probabilistic Formulation}\label{sec:formulation}

We formulate the task of clothing co-parsing as a probabilistic model.
Let ${\bf I}=\{I_i\}_{i=1}^N$  denote a set of clothing images with tags $\{T_i\}_{i=1}^N$. Each image $I$  is represented by a set of superpixels $I=\{s_j\}_{j=1}^M$, which will be further grouped into several coherent regions under the guidance of segmentation propagation. Each image $I$ is associated with four additional variables:
\begin{myitemize}
  \item[a)] the regions $\{r_k\}_{k=1}^K$, each of which is consisted of a set of superpixels;
  \item[b)] the garment label for each region: $\ell_k\in T, k=1, ... , K$ ;
  \item[c)] the E-SVM weights $w_{k}$ trained for each selected region;
  \item[d)] the segmentation propagations $C=(x, y, m)$, where $(x, y)$ is the location and $m$ is the segmentation mask of an E-SVM, indicating segmentation mask $m$ can be propagated to the position $(x, y)$ of $I$, as illustrated in Figure~\ref{fig:framework} (a).
\end{myitemize}\vspace{0.3em}

Let ${\bf R}=\{R_i=\{r_{ik}\}\}$, ${\bf L}=\{L_i=\{\ell_{ik}\}\}$, ${\bf W}=\{W_i=\{w_{ik}\}\}$ and ${\bf C}=\{C_i\}$. We optimize the parameters by maximizing the following posterior probability:

\begin{small}
\begin{equation}\label{eq:obj}
\begin{split}
\{ {\bf L}^\ast, {\bf R}^\ast, {\bf W}^\ast, {\bf C}^\ast\}=
\arg\max P(  {\bf L}, {\bf R}, {\bf W}, {\bf C} | {\bf I}),
\end{split}
\end{equation}
\end{small}
which can be factorized as
\begin{small}
\begin{equation}\label{eq:factor}
\begin{split}
P( {\bf L}, {\bf R}, {\bf W}, {\bf C}\} | {\bf I}) &\propto \overbrace{P({\bf L}|{\bf R},{\bf C})}^{co-labeling}\times\\
&\overbrace{\prod_{i=1}^N P(R_i| C_i, I_i) P(W_i | R_i)\times P(C_i|{\bf W},I_i)}^{co-segmentation} .
\end{split}
\end{equation}
\end{small}
The optimization of Eq. (\ref{eq:factor}) includes two phases: (I) clothing image co-segmentation and (II) co-labeling.

In {\em phase} (I), we obtain the optimal regions by maximizing $P(R|C, I)$ in Eq. (\ref{eq:factor}). We introduce the superpixel grouping indicator $o_j\in\{1, \cdots, K\}$, which indicates to which of the $K$ regions the superpixel $s_j$ belongs. Then each region can be denoted as a set of superpixels, as $r_k = \{s_j | o_j=k\}$. Given the current segmentation propagation $C$, $P(R|C, I)$ can be defined as,
\begin{small}
\begin{equation}\label{eq:superpixel_grouping}
\begin{split}
P(R|C, I) &= \prod_k P(r_k|C, I) \propto \prod_j P(o_j | C, I) \\
&\propto \prod_{j=1}^M P( o_j, s_j) \prod_{mn} P(o_m, o_n, s_m, s_n|C),
\end{split}
\end{equation}
\end{small}
where the unary potential $P( o_j, s_j) \propto \exp \{-d(s_j, o_j)\}$  indicates the probability of superpixel $s_j$ belongs to a region, where $d(s_j, o_j)$ evaluates the spatial distance between $s_j$ and its corresponding region.
$P(o_m, o_n, s_m, s_n|C) $ is the pairwise potential function, which encourages smoothness between neighboring superpixels.

After grouping superpixels into regions, we then select several coherent regions to train an ensemble of E-SVMs, by maximizing $P(W|R)$ defined as follows,
\begin{equation}\label{eq:esvm}
\begin{split}
    P(W|R)&=  \prod_k P(w_k|r_k) \propto  \prod_k \exp\{-E(w_k, r_k)\cdot \phi(r_j) \},
\end{split}
\end{equation}
where $ \phi(r_j)$ is an indicator exihibiting whether $r_j$ has been chosen for training E-SVM. $E(w_k, r_k)$ is the convex energy function of E-SVM.

Finally,  we define $P(C_i|{\bf W},I_i)$ in Eq. (\ref{eq:factor}) based on the responses of E-SVM classifiers. This probability is maximized by selecting the top $k$ detections of each E-SVM classifier as the segmentation propagations by the sliding window scheme.

%

In {\em phase} (II), we assign a garment tag to each region by modeling the problem as a multi-image graphical model,

\begin{equation}\label{eq:co-lab}
\begin{split}
P({\bf L}|{\bf R},{\bf C})  \propto &\prod_i^N \prod_k^K P(\ell_{ik}, r_{ik}) \\
&  \prod_{mn}P(\ell_m, \ell_n, r_m, r_n) \prod_{uv} Q(\ell_u, \ell_v, r_u, r_v|\mathbf{C}),
\end{split}
\end{equation}
where $P(\ell_{ik}, r_{ik})$ represents the singleton potential of assigning label $\ell_{ik}$ to region $r_{ik}$, and  $P(\ell_m, \ell_n, r_m, r_n)$ the interior affinity model capturing compatibility among regions within one image, and $Q(\ell_u, \ell_v, r_u, r_v|\mathbf{C})$ the exterior affinity model for regions belonging to different images, in which $r_u$ and $r_v$ are connected under the segmentation propagation $\mathbf{C}$. Details are discussed in Section~\ref{sec:co-label}.

%

\section{Clothing Co-Parsing}\label{sec:co-parsing}
In this section, we describe the two phases of clothing co-parsing, including their implementation details. The overall procesure is outlined in Algorithm \ref{alg:co-parsing}.

\subsection{Unsupervised Image Co-Segmentation}\label{sec:co-seg}
The optimization in the co-segmentation is to estimate a variable while keeping others fixed, \eg, estimating $R$, with $W, C$ fixed. Thus the first phase iterates between three steps:

\noindent\textbf{i. Superpixel Grouping}: The MRF model defined in Eq. (\ref{eq:superpixel_grouping}) is the standard pipeline for superpixel grouping. However, the number of regions need to be specified, which is not an applicable assumption of our problem.

To automatically determine the number of regions, we replace the superpixel indicator $o_j$ by a set of \emph{binary} variables $o^e$ defined on the edges between neighboring superpixels. Let $e$ denote an edge, $o^e=1$ if two superpixels $s^e_1$ and $s^e_2$ connected by $e$ belong to the same region, otherwise $o^e=0$. We also introduce a binary variable $o^c $ with $o^c=1$ indicating all the superpixels covered by the mask of the segmentation propagation $c$ belong to the same region, otherwise $o^c=0$. Then maximizing Eq. (\ref{eq:superpixel_grouping}) is equivalent to the following linear programming problem,
\begin{equation}\label{eq:superpixel_grouping_lp}
\begin{split}
&\arg\min_{o^e, o^c} \sum_{e} d(s^e_1, s^e_2)\cdot o^e - \sum_{ c\in C} h(\{s_j|s_j\subset c\})\cdot o^c,
\end{split}
\end{equation}
where $d(s^e_1,s^e_2)$ is the dissimilarity between two superpixels, and $h(\cdot)$ measures the consistence of grouping all superpixels covered by an E-SVM mask into one region.
Eq. (\ref{eq:superpixel_grouping_lp}) can be efficiently solved via the cutting plane algorithm as introduced in~\cite{solution}.

 The dissimilarity $d(s^e_1,s^e_2)$ in Eq. (\ref{eq:superpixel_grouping_lp}) is defined together with the detected contours: $d(s^e_1,s^e_2) = 1$  if there exists any contour across the area covered by $s^e_1$ and $s^e_2$, otherwise $d(s^e_1,s^e_2)  = 0$. $h(\{s_j|s_j\subset c\})$ is defined as the normalized total area of the superpixels covered by the template $c$.

\vspace{0.5em}\noindent\textbf{ii. Training E-SVMs}: The energy $E(w_k, r_k)$ introduced in Eq. (\ref{eq:esvm}) is the convex energy function of E-SVM as follows,
{\small
\begin{equation}\label{eq:svm_obj}
\frac{1}{2}||w_k||^2\!+\!\lambda_1\!\!\max(0,1\!-\!w_k^Tf(r_k))\!+\!\lambda_2\!\!\sum_{r_n\in N_E}\max\!(0,1\!-\!w_k^Tf(r_n)).
\end{equation}
}
$N_E$ denotes the negative examples, and $f(\cdot)$ is the feature of a region, following \cite{ESVM}. $\lambda_1$ and $\lambda_2$ are two regularization parameters. Thus maximizing $P(W|R)$ is equivalent to minimizing the energy in Eq. (\ref{eq:svm_obj}), \ie, training the parameters of the E-SVM classifiers by the gradient descent.

We train an E-SVM classifier for each of the selected regions: each selected region is considered as a positive example (exemplar), and a number of patches outside the selected region are croped as negative examples. In the implementation, we use HOG as the feature for each region. The region selection indicator $\phi(r_j)$ in Eq. (\ref{eq:esvm}) is determined by the automated  saliency detection~\cite{PISA2013}.  For computational efficiency, we only train E-SVMs for high confident foreground regions, \ie  regions containing garment items.

\vspace{0.5em}\noindent\textbf{iii. Segmentation Propagation}: We search for possible propagations by sliding window method. However, as each E-SVM is trained independently, their responses may not be compatible. We thus perform the calibration by  fitting a logistic distribution with parameters $\alpha_E$ and $\beta_E$ on the training set.  Then the E-SVM response can be  defined as,
\begin{equation}\label{eq:esvm_score}
S_E(f; w) = \frac{1}{1+ \exp(-\alpha_E (w^Tf-\beta_E)},
\end{equation}
where $f$ is the feature vector of the image patch covered by the sliding window.




\subsection{Contextualized Co-Labeling}\label{sec:co-label}

\begin{figure}[tpb]
\begin{center}
   \epsfig{figure=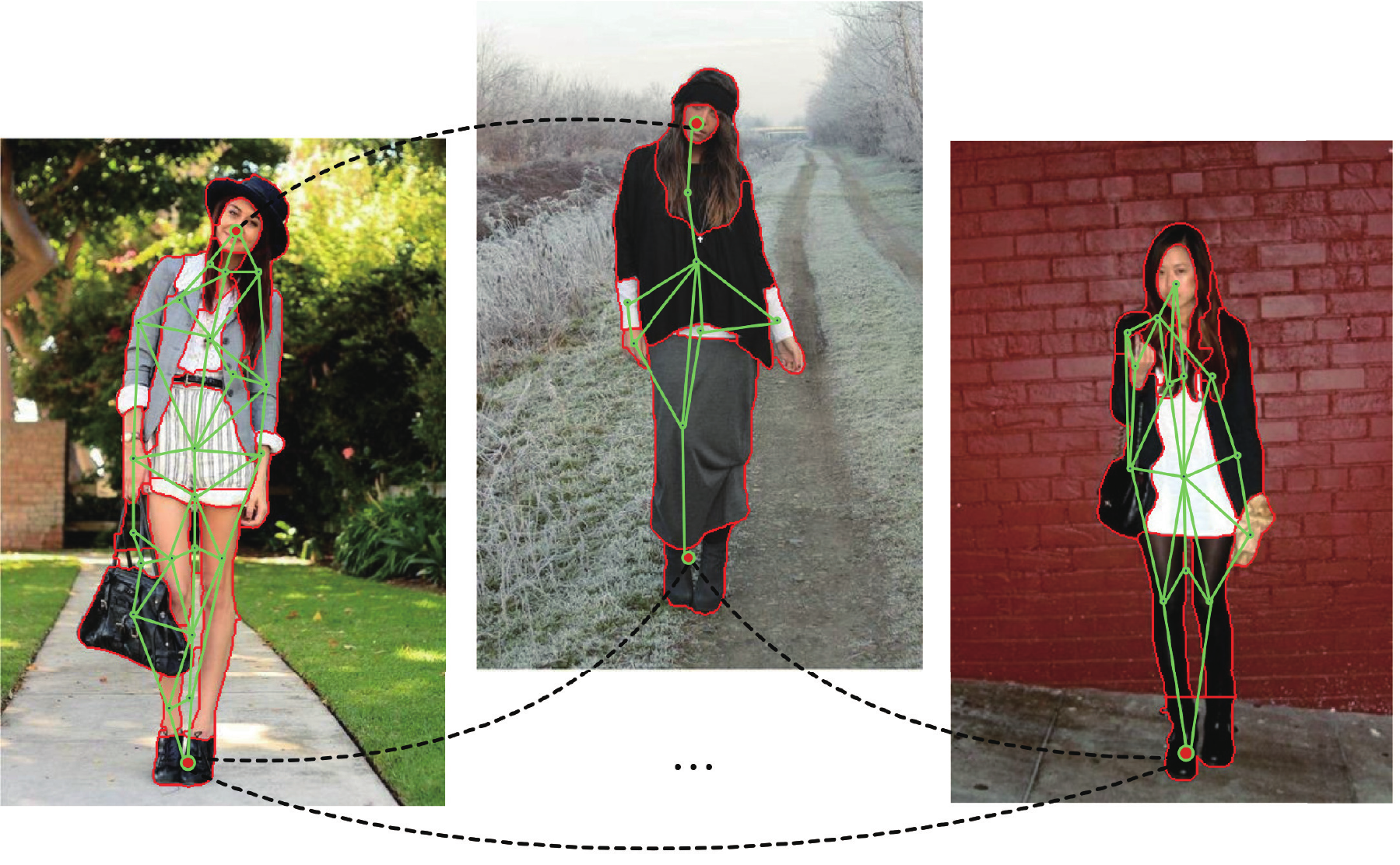, width=1\linewidth}
\end{center}
\vspace{-3mm}
   \caption { We perform co-labeling by optimizing a multi-image graphical model, \ie an MRF connecting all the images in the database. A toy example of the model is illustrated above, where the green solid lines are interior edges between adjacent regions within the same images while the black dashed lines are exterior edges across different images. Note that the connections among different images are determined by the segmentation propagation.}
   \label{fig:inference}
\end{figure}

In this phase, each image is represented by a set of coherent regions, and we assign a garment tag to each region by optimizing a multi-image graphical model , \ie an MRF connecting all the images in the database, which is defined in Eq. (\ref{eq:co-lab}). We define two types of edges on the graph:  the interior edges connecting neighboring regions within an image, and the exterior edges connecting regions of different images matched by the propagated segmentation.  A toy example of the graphical model  is showed in Figure ~\ref{fig:inference}. Specifically, two types of clothing contexts are exploited providing informative constraints during inference.

The  {\bf singleton potential} $P(\ell_k, r_k)$  defined  in Eq. (\ref{eq:co-lab}) incorporates a region appearance model with the garment item location context.

For each type of garment, we train the appearance model as an SVM classifier based on local region appearance. Let $S(f(r_k), \ell_k)$ denote the score of the appearance model, and $f(r_k)$ a feature vector of $40$-bins concatenated by the color and gradient histograms. We define the potential of assigning label $\ell_k$ to region $r_k$ as,
\begin{equation}\label{eq:unarybp}
P(\ell_k, r_k) = \mathrm{sig}(S(f(r_k), \ell_k))\cdot G_{\ell_k}(X_k),
\end{equation}
where $\mathrm{sig}(\cdot)$ indicates the sigmoid function, and $X_k$ the center of region $r_k$. The location context $G_{\ell_k}(X_k)$ is defined upon the 2-D Gaussian distribution as,
\begin{equation}\label{eq:garment_location}
G_{\ell_k}(X_k) \sim \mathcal{N}(\mu_{\ell_k}, \Sigma_{\ell_k}),
\end{equation}
where $\mu_{\ell_k}$ and $\Sigma_{\ell_k}$ represent the mean and the covariance of the location of garment item $\ell_k$, respectively, which can be estimated over the training set.

The {\bf interior affinity} $P(\ell_m, \ell_n, r_m, r_n)$  in Eq. (\ref{eq:co-lab}) of two adjacent regions $r_m$ and $r_n$ is defined on two terms within an image: their appearance compatibility and mutual interactions, as
\begin{equation}
P(\ell_m, \ell_n, r_m, r_n)= \Phi(\ell_m, \ell_n, r_m, r_n) \Psi(\ell_m, \ell_n).
\end{equation}
The appearance compatibility function $\Phi(\ell_m, \ell_n, r_m, r_n)$ encourages regions with similar appearance to have the same tag:
\begin{equation}\label{eq:pairbp}
 \Phi(\ell_m, \ell_n, r_m, r_n) = \exp\{- \textbf{1}(\ell_m = \ell_n) d(r_m, r_n) \}
\end{equation}
where $\textbf{1}(\cdot)$ is the indicator function, and $d(r_m, r_n)$ is the $\mathcal{X}^2$-distance between the appearance feature of two regions.

$\Psi(\ell_m, \ell_n)$ models the mutual interactions of two different garment items $\ell_m$ and $\ell_n$. This term is simple but effective, since some garments are likely to appear as neighbors in an image, \eg $coat$ and $pants$, while others are not, \eg $hat$ and $shoes$. In practice, we calculate $\Psi(\ell_m, \ell_n)$ by accumulating the frequency of they appearing as neighbors over all adjacent image patches in the training data.

The {\bf exterior affinity} $Q(\ell_u, \ell_v, r_u, r_v|\mathbf{C})$ of  Eq. (\ref{eq:co-lab}) across different images constrains that regions in different images sharing similar appearance and locations should have high probability to have the same garment tag. We thus have,
{\small
\begin{equation}
Q(\ell_u, \ell_v, r_u, r_v | \mathbf{C}) =G_{\ell_u}(X_u) G_{\ell_v}(X_v)\Phi(\ell_u, \ell_v, r_u, r_v) ,
\end{equation}
}
in which the terms were clearly defined in Eq.(\ref{eq:garment_location}) and Eq.(\ref{eq:pairbp}). Finally, we adopt the Graph Cuts to optimize the multi-image graphical model.

\begin{algorithm}
\caption{The Sketch of Clothing Co-Parsing}
\label{alg:co-parsing}
{
\begin{algorithmic}
\REQUIRE ~~\\
A set of clothing images ${\bf I}=\{I_i\}_{i=1}^N$  with tags $\{T_i\}_{i=1}^N$.  \\

\ENSURE ~~\\                           
The segmented regions ${\bf R}$ with their corresponding labels ${\bf L}$. \\
\vspace{0.5em}
\PHASE \emph{(I):  Image Co-Segmentation}
\MYREPEATE
    \STATE
    \begin{myitemize}
      \item[1] For each image $I$, group its superpixels into regions $R$ under the guidance of the segmentation propagations $C$ by maximizing $P( R | C, I)$ in Eq. (\ref{eq:superpixel_grouping});
      \item[2] Train E-SVM parameters for each selected region by minimizing the energy in Eq. (\ref{eq:svm_obj}).
      \item[3] Propagate segmentations across images by detections from the trained E-SVM classifiers by Eq. (\ref{eq:esvm_score}).
    \end{myitemize}
\MYUNTIL {Regions are not changed during the last iteration}

\vspace{0.5em}
\PHASE \emph{ (II):  Contextualized Co-Labeling}
\begin{myitemize}
      \item[1] Construct the multi-image graphical model;
      \item[2] Solving the optimal label assignment ${\bf L}^*$ by optimizing the probability defined on the graphical model as in Eq. (\ref{eq:co-lab})  by Graph Cuts.
    \end{myitemize}
\end{algorithmic}
}
\end{algorithm}

\section{Experiments}\label{sec:experiments}
We first introduce the clothing parsing datasets, and present the quantitative results and comparisons. Some qualitative results are exhibited as well.

{\em Parameter settings:} We use gPb contour detector~\cite{arbelaez2011contour}  to obtain superpixels and contours, and the threshold of the detector  is adapted to obtain about 500 superpixels for each image. Contours help define $d(s^e_1, s^e_2)$ in Eq. (\ref{eq:superpixel_grouping_lp}) were obtained by setting the threshold to 0.2. For training E-SVMs, we set  $\lambda_1 = 0.5$ and $\lambda_2 = 0.01$ in Eq. (\ref{eq:svm_obj}) to train E-SVMs. The appearance model in Sec. \ref{sec:co-label} is trained by a multi-class SVM using one-against-one decomposition with an Gaussian kernel.

\subsection{Clothing Image Datasets}\label{sec:datasets}
\begin{figure*}[t]
\begin{center}
   \epsfig{figure=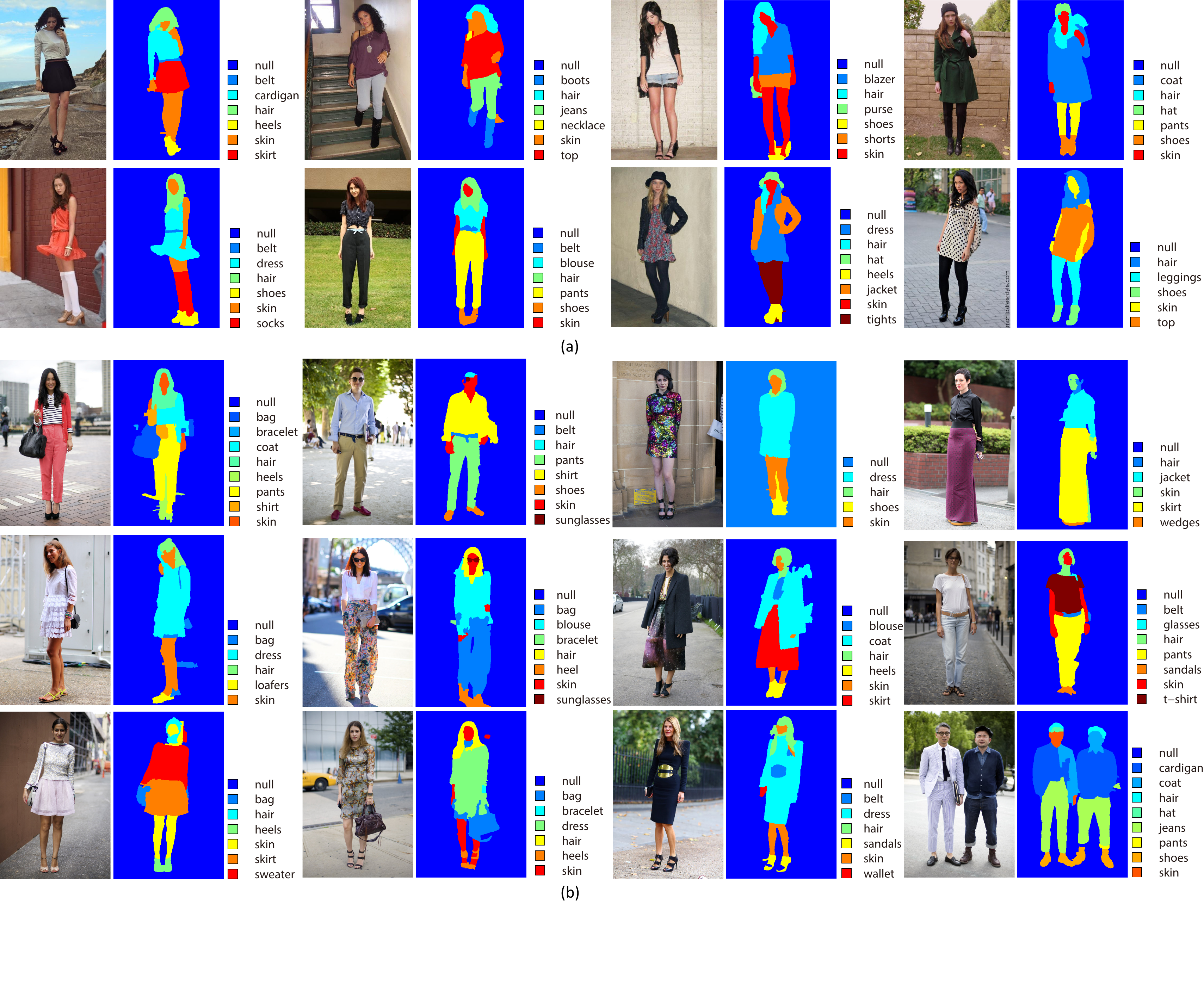, width=1\textwidth}
\end{center}
   \vspace{-15mm}\caption{Some successful parsing results on (a) Fashionista (b) CCP.}
   \label{fig:good-result}
\end{figure*}

We evaluate our framework on two datasets: Clothing Co-Parsing\footnote{http://vision.sysu.edu.cn/projects/clothing-co-parsing/} (CCP) and Fashionista~\cite{Fashion}. CCP is created by us, consisting of $2,098$ high-resolution fashion photos with huge human/clothing variations, which are in a wide range of styles, accessories, garments, and poses. More than 1000 images of CCP are with superpixel-level annotations with totally 57 tags, and the rest of images are annotated with image-level tags. All annotations are produced by a professional team. Some examples of CCP are shown in Figure~\ref{fig:good-result}. Fashionista contains $158,235$ fashion photos from fashion blogs which are further separated into an annotated subset containing $685$ images with superpixel-level ground truth, and an unannotated subset associated with possibly noisy and incomplete tags provided by the bloggers. The annotated subset of Fashionista contains $56$ labels, and some garments with high occurrences in the dataset are shown in Figure \ref{fig:recall}.

\subsection{Quantitative Evaluation}\label{sec:quantitative}

To evaluate the effectiveness of our framework, we compare our method with three state-of-arts: (1) PECS~\cite{Fashion} which is a fully supervised clothing parsing algorithm that combines \underline{p}ose \underline{e}stimation and \underline{c}lothing \underline{s}egmentation, (2) the \underline{b}i-layer \underline{s}parse \underline{c}oding (BSC)~\cite{liu2009label}  for uncovering the label for each image region, and (3) the \underline{s}emantic \underline{t}exton \underline{f}orest (STF)~\cite{shotton2008semantic}, a standard pipeline for semantic labeling.

The experiment is conducted both on Fashionista and CCP datasets. Following the protocol in~\cite{Fashion}, all measurements use 10-fold cross validation, thus 9 folds for training as well as for tuning free parameters, and the remaining for testing.
The performances are measured by average Pixel Accuracy (aPA) and mean Average Garment Recall (mAGR), as in \cite{Fashion}. As \emph{background} is the most frequent label appearing in the datasets, simply assigning all regions to be \emph{background} achieves $77.63\%$ / $77.60\%$ accuracy,  and $9.03 \%$ / $15.07\%$ mAGR,  on the Fashionista and the CCP dataset respectively. We treat them as the baseline results.

\begin{figure}[tpb]
\begin{center}
   \epsfig{figure=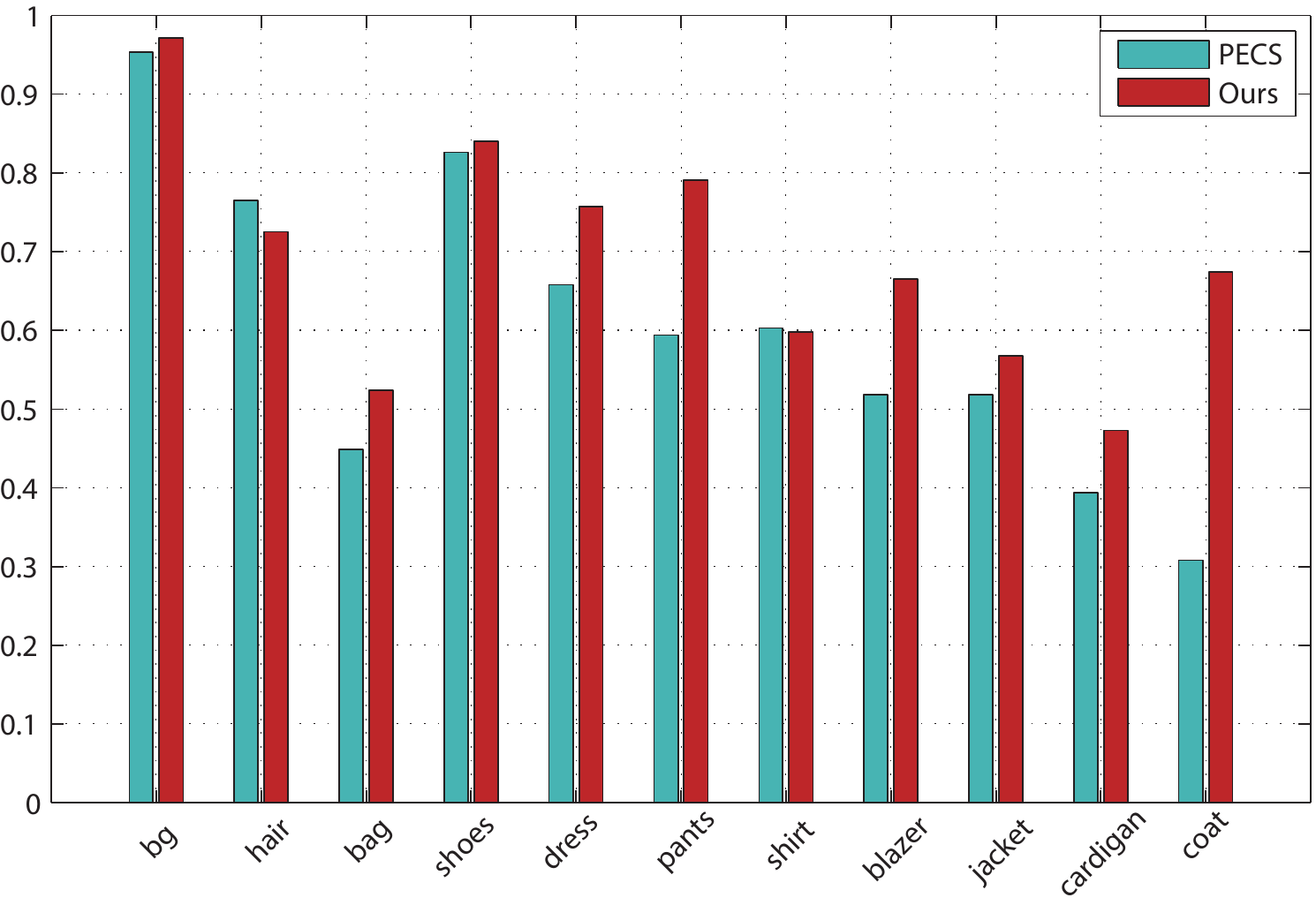, width=0.8\linewidth}
\end{center}
\vspace{-5mm}
   \caption { Average recall of some garment items with high occurrences in Fashionista. }
   \label{fig:recall}
\end{figure}

Table \ref{tab:fashionista-ccp} reports the clothing parsing performance of each method on the Fashionista and CCP datasets. On both datasets, our method achieves much superior performances over the BSC and STF methods, as they did not address the specific clothing knowledges. We also outperform the state-of-the-art clothing parsing system PECS  on both datasets. As images of the CCP database include more complex backgrounds and clothing styles, the advantage of our approach is better demonstrated. In fact, the process of iterative image co-segmentation effectively suppresses the image clutters and generates coherent regions, and the co-labeling phase handles better the variants of clothing styles by incorporating rich priors and contexts. In addition, we report the average recall of several frequently occurring garment items in Fashionista dataset in Figure \ref{fig:recall}.


\begin{table}[tpb]
\begin{small}
\begin{center}
\begin{tabular*}{1\linewidth}{@{\extracolsep{\fill}}l  c  c | c c}
\hline
& \multicolumn{2}{c|}{Fashionista} & \multicolumn{2}{c}{CCP} \\
\hline
Methods & aPA & mAGR & aPA & mAGR \\
\hline
Ours-full &\textbf{90.29} & \textbf{65.52} & \textbf{88.23} & \textbf{63.89}\\
PECS~\cite{Fashion} & 89.00 & 64.37 & 85.97 & 51.25\\
 BSC~\cite{liu2009label} & 82.34 & 33.63 & 81.61 & 38.75\\
STF~\cite{shotton2008semantic} & 68.02 &  43.62 &  66.85 &  40.70 \\
\hline
Ours-1 & 89.69 &  61.26 &  87.12 &  61.22 \\
Ours-2 & 88.55  &  61.13 &  86.75 &  59.80 \\
Ours-3 & 84.44 &  47.16 &  85.43 &  42.50 \\
\hline
Baseline &77.63 & 9.03  &  77.60 & 15.07\\

\hline
\end{tabular*}
\center {\caption{\footnotesize Clothing parsing results (\%) on the Fashionista and the CCP dataset. As \emph{background} is the most frequent label appears in the datasets, assigning all regions to be \emph{background} is adopted our baseline comparison. We compare our full system (Ours-full) to three state-of-the-art methods including PECS~\cite{Fashion}, BSC~\cite{liu2009label}, and  STF~\cite{shotton2008semantic}.
We also present an empirical analysis to evaluate the effectiveness of the main components of our system. Ours-1 and Ours-2  evaluate the effectiveness of the co-labeling phase by only employing the exterior affinity, and by only using the interior affinity, respectively. Ours-3 evaluates the performance of superpixel grouping in the co-segmentation phase. }\label{tab:fashionista-ccp}}
\end{center}
\end{small}
\end{table}

\vspace{0.3em}\noindent\textbf{Evaluation of Components.}
We also present an empirical analysis to demonstrate the effectiveness of the main components of our system. Ours-1 and Ours-2 in Table  \ref{tab:fashionista-ccp} evaluate the effectiveness of the co-labeling phase by only employing the exterior affinity, and by only using the interior affinity, respectively. Ours-3 evaluates the performance of superpixel grouping in the co-segmentation phase. Ours-1 achieves the best result compared to Ours-2 and Ours-3 due to the importance of mutual interactions between garment items, thus performing co-labeling on a multi-image graphical model benefits the clothing parsing problem.



\subsection{Qualitative Evaluation}\label{sec:qualitative}

  Figure \ref{fig:good-result} illustrates some successful parsing results for exemplary images from both Fashionista and CCP. Our method is able to  parse clothes accurately even in some challenging illumination and complex background conditions ({r1c2}\footnote{We use ``r1c1" to denote the image in row 1, column 1. }, {r4c2}). Moreover, our method could even parse some small garments such as \emph{belt} ({r1c1, r2c1, r2c2, r3c2}), \emph{purse} ({r1c3}), \emph{hat} ({r1c4, r2c3}), and \emph{sunglasses} ({r4c2}). For reasonably ambiguous clothing patterns such as dotted t-shirt or colorful dress, our framework could give satisfying results ({r2c4, r5c2}). In addition, the proposed method could even parse several persons in a single image simultaneously ({r5c5}).

\begin{figure*}[!ht]
\begin{center}
    \epsfig{figure=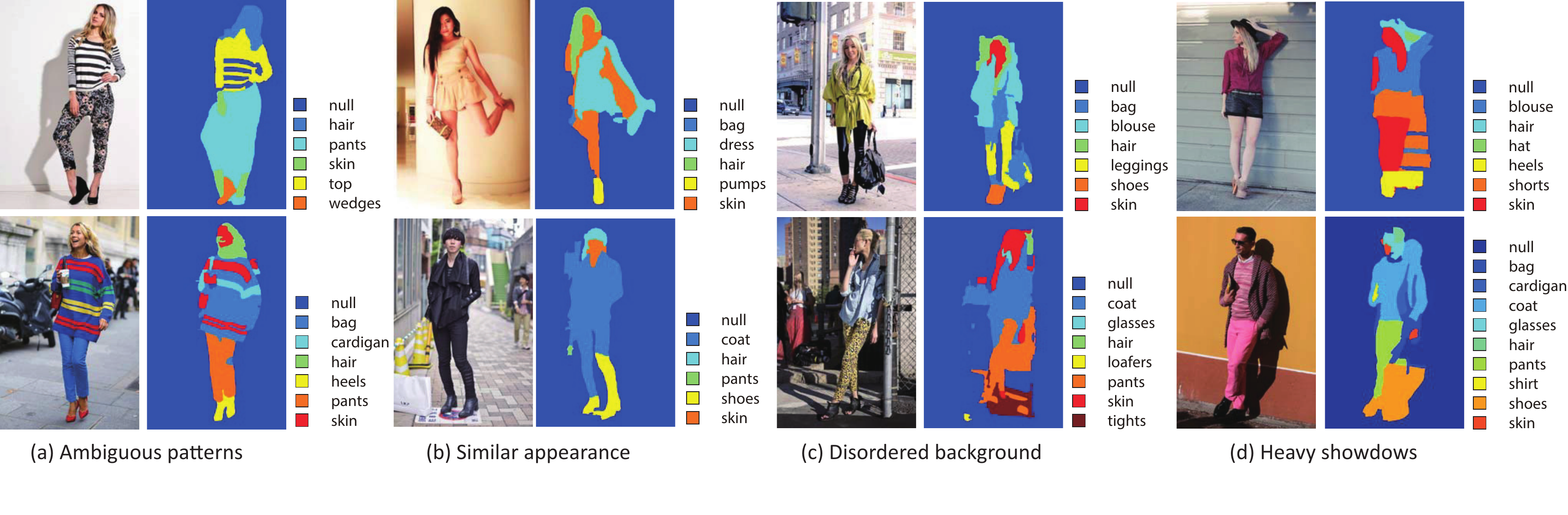, width=1\textwidth}
\end{center}
   \vspace{-10mm}\caption{Some failure cases on Fashionista (1st row) and CCP (2nd row).}
   \label{fig:bad-result}
\end{figure*}

 Some failure cases are shown in Figure \ref{fig:bad-result}. Our co-parsing framework may lead wrong results under following scenarios: (a) ambiguous patterns exist within a clogging garment item; (b) different  clothing garment items share similar appearance; (c) background is extremely disordered; (d) illumination condition is poor.

\subsection{Efficiency}
All the experiments are carried out on an Intel Dual-Core E6500 (2.93 GHz) CPU and 8GB RAM PC. The run-time complexity of the co-segmentation phase scales linearly with the number of iterations, and each iteration costs about 10 sec per image. The co-labeling phase costs less than 1 minute to assign labels to a database of 70 images, which is very effective due to the consistent regions obtained from the co-segmentation phase. And the Graph Cuts algorithm converges in 3-4 iterations in our experiment.

\section{Conclusions}\label{sec:conclusions}

This paper has proposed a well-engineered framework for joint parsing a batch of clothing images given the image-level clothing tags. Another contribution of this paper is a high-resolution street fashion photos dataset with annotations. The experiments demonstrate that our framework is effective and applicable compared with the state-of-the-art methods. In future work, we plan to improve the inference by iterating the two phases to bootstrap the results. In addition, the parallel implementation would be studied to adapt the large scale applications.

{\small
\bibliographystyle{ieee}
\bibliography{mybib}
}
\end{document}